\documentclass[10pt,twocolumn,letterpaper]{article}

\usepackage{cvpr}
\usepackage{times}
\usepackage{epsfig}
\usepackage{graphicx}
\usepackage{amsmath}
\usepackage{amssymb}
\usepackage{booktabs}
\usepackage{threeparttable}
\usepackage{amsmath,amssymb}
\usepackage{helvet}  
\usepackage{courier}  
\usepackage{url}  
\usepackage{graphicx}  
\usepackage{multirow}
\usepackage{array}
\usepackage{float} 
\usepackage{mwe}

\usepackage[breaklinks=true,bookmarks=false]{hyperref}

\cvprfinalcopy 

\def\cvprPaperID{5442} 

\ifcvprfinal\fi

\title{TACNet: Transition-Aware Context Network for \\ 
	Spatio-Temporal Action Detection}

\author{Lin Song$^{1}$\thanks{indicates equal contribution.} \quad Shiwei Zhang$^{2}$\footnotemark[1] \quad Gang Yu$^3$ \quad Hongbin Sun$^1$\thanks{indicates corresponding author.} \\
$^1$ Institute of Artificial Intelligence and Robotics, Xi'an Jiaotong Univeristy. \\
$^2$ Artificial Intelligence and Automation, Huazhong University of Science and Technology. \\
$^3$ Megvii Inc. (Face++).\\
\{stevengrove@stu, hsun@mail\}.xjtu.edu.cn, ~swzhang@hust.edu.cn, ~yugang@megvii.com}

\makeatletter
\let\@oldmaketitle\@maketitle
\renewcommand{\@maketitle}{\@oldmaketitle
\includegraphics[width=\textwidth]{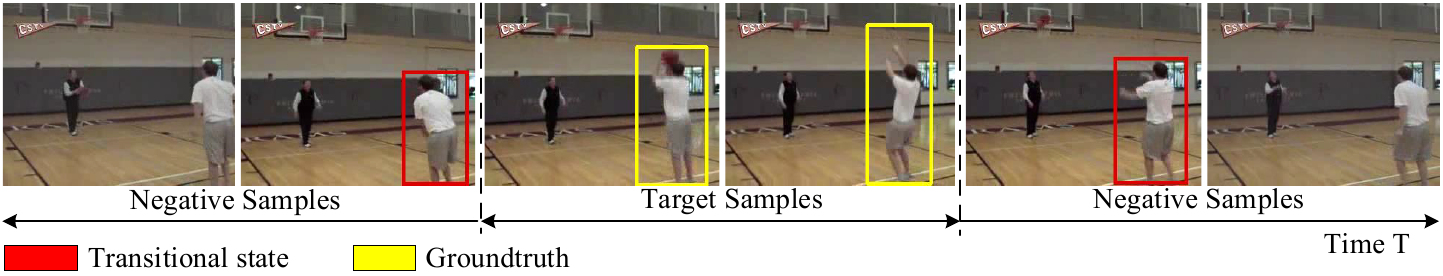}
{Figure 1: Diagram of transitional state. There are some ambiguous states around but not belong to the target actions, and it is hard to distinguish them. We define the states as ``transitional state'' (red boxes). 
If we can effectively distinguish these states, we can improve the ability of temporal extent detection.}\bigskip}
\label{fig:motivation}
\makeatother
\begin{document}\maketitle 

\thispagestyle{empty}



\begin{abstract}
Current state-of-the-art approaches for spatio-temporal action detection have achieved impressive results but remain unsatisfactory for temporal extent detection. 
The main reason comes from that, there are some ambiguous states similar to the real actions which may be treated as target actions even by a well-trained network.
In this paper, we define these ambiguous samples as ``transitional states'', and propose a Transition-Aware Context Network (TACNet) to distinguish transitional states. 
The proposed TACNet includes two main components, i.e., temporal context detector and transition-aware classifier. 
The temporal context detector can extract long-term context information with constant time complexity by constructing a recurrent network.
The transition-aware classifier can further distinguish transitional states by classifying action and transitional states simultaneously.
Therefore, the proposed TACNet can substantially improve the performance of spatio-temporal action detection.
We extensively evaluate the proposed TACNet on UCF101-24 and J-HMDB datasets. 
The experimental results demonstrate that TACNet obtains competitive performance on JHMDB and significantly outperforms the state-of-the-art methods on the untrimmed UCF101-24 in terms of both frame-mAP and video-mAP.

\end{abstract}

\section{Introduction}\label{sec:intro}

Action detection focuses both on classifying the actions present in a video and on localizing them in space and time.
It has been receiving more and more attention from researchers because of its various applications. Action detection has already served as a critical technology in anomaly detection, human-machine interaction, video monitoring, etc. 
Currently, most of action detection approaches~\cite{hou2017tube,peng2016multi,singh2017online,2015fast} separate the spatio-temporal detection into two stages, i.e., spatial detection and temporal detection. These approaches adopt the detectors based on deep neural networks~\cite{girshick_fast_2015,liu_ssd:_2016} to spatially detect action in the frame level. Then, they construct temporal detection by linking frame-level detections and applying some objective functions, such as maximum subarray method~\cite{peng2016multi}, to create spatio-temporal action tubes. Because these methods treat video frames as a set of independent images, they can not exploit the temporal continuity of videos. Thus, their detection results are actually unsatisfactory.

To address this issue, ACT~\cite{kalogeiton2017action} employs a stacked strategy to exploit the short-term temporal continuity for clip-level detection and significantly improves the performance of spatio-temporal action detection. 
However, ACT still can not extract long-term temporal context information, which is critical to the detection of many action instances, such as ``long jump''. 
Moreover, due to the two separate stages in action detection, ACT can not thoroughly address the time error induced by ambiguous samples, which are illustrated as red boxes in Figure 1. 
In this paper, the ambiguous sample is defined as ``transitional state'', which is close to the action duration but does not belong to the action. 
According to the error analysis of ACT detector, 35\%-40\% of the total errors~\cite{kalogeiton2017action} are time errors, which are mainly caused by transitional states.
Therefore, to further improve the performance of spatio-temporal action detection, it is critical to extract long-term context information and distinguish the transitional states in a video sequence.

The above observations motivate this work. In particular, we propose a Transition-Aware Context Network (TACNet) to improve the performance of spatio-temporal action detection. 
The proposed TACNet includes two main components, i.e., temporal context detector and transition-aware classifier. 
The temporal context detector is designed based on standard SSD framework, but can encode the long-term context information by embedding several multi-scale Bi-directional Conv-LSTM \cite{li2018videolstm} units. 
To the best of our knowledge, this is the first work to combine Conv-LSTM with SSD to construct a recurrent detector for action detection. 
The transition-aware classifier can distinguish transitional states by classifying action and action states simultaneously. More importantly, we further propose a common and differential mode scheme to accelerate the convergence of TACNet.
Therefore, the proposed TACNet can not only extract long-term temporal context information but also distinguish the transitional states.
We test the proposed TACNet on UCF101-24~\cite{soomro2012ucf101} and J-HMDB~\cite{jhuang2013towards} datasets and achieve a remarkable improvement 
in terms of both frame- and video-level metrics on both datasets.

In summary, we make the following three contributions: 
\begin{itemize}
\item we propose a temporal context detector to extract long-term temporal context information efficiently with constant time complexity;
\item we design a transition-aware classifier, which can distinguish transitional states and alleviate the temporal error of spatio-temporal action detection;
\item we extensively evaluate our TACNet in untrimmed videos from the UCF-24 dataset and achieve state-of-the-art performance.
\end{itemize}

\section{Related Work}
\label{sec:relatedwork}

Spatio-temporal action detection methods can generally be classiﬁed into two categories: weakly and fully supervised methods. Although we concentrate on fully supervised methods in this paper, weakly supervised methods have also achieved signiﬁcant improvement in recent years.
The purpose of these methods is to detect actions only with video-level labels but without frame-level bounding box annotations. These methods can significantly reduce annotation costs and are more suitable to process large unannotated video data. 
Multi-Instance Learning (MIL) is one of the frequently-used approaches for weakly supervised spatio-temporal action detection.
In ~\cite{2011weakly}, Siva \emph{et al.} transforms the weakly supervised action detection as a MIL problem. They globally optimize both inter- and intra-class distances to locate interested actions. Multi-fold MIL scheme is then proposed in ~\cite{2014multi} to prevent training from prematurely locking onto erroneous object detection.
Recently, deep model and attention mechanisms are also employed in deep model based weakly supervised methods. 
The methods in ~\cite{2015action,li2018videolstm} apply attention mechanism to focus on key volumes for action detection. 
Besides, Mettes~\emph{et al.} ~\cite{2016spot,2017localizing} proposes to apply point annotations to perform action detection.

Compared with weakly supervised methods, fully supervised methods can leverage bounding box level annotations to achieve remarkable performance for spatio-temporal action detection. Many approaches are proposed to construct action tubes.
Gkioxari \emph{et al.}~\cite{gkioxari2015finding} firstly proposes to apply the linking algorithm on frame-level detection to generate action tubes. 
Peng \textit{et al.}~\cite{peng2016multi} improves frame-level action detection by stacking optical flow over several frames and also proposes a maximum subarray method for temporal detection.
Weinzaepfel \textit{et al.}~\cite{weinzaepfel2015learning} improves the linking algorithm by using a tracking-by-detection method.
Singh \textit{et al.}~\cite{singh2017online} designs online algorithm to incrementally generate action tubes for real-time action detection. However, these methods do not explore the temporal information of actions, and the performance is still unsatisfactory.
To encode the temporal information, Saha \textit{et al.}~\cite{saha2017amtnet} and Hou \textit{et al.}~\cite{hou2017tube} extend classical region proposal network (RPN) to 3D RPN, which generates 3D region proposals spanned by several successive video frames. Becattini \textit{et al.}~\cite{becattini2017done} adopts LSTM to predict action progress. 
Zhu \textit{et al.}~\cite{zhu2017tornado} proposes a two-stream regression network to generate temporal proposals. 
In contrast, Kalogeiton \emph{et al.}~\cite{kalogeiton2017action} stacks the feature maps of multiple successive frames by SSD detector to predict score and regression on anchor cuboid and achieves the state-of-the-art performance. 
Therefore, this paper uses ACT~\cite{kalogeiton2017action} as the baseline to compare and evaluate the action detection performance of TACNet.

\addtocounter{figure}{1}
\begin{figure*}
\centering
\includegraphics[width=18cm]{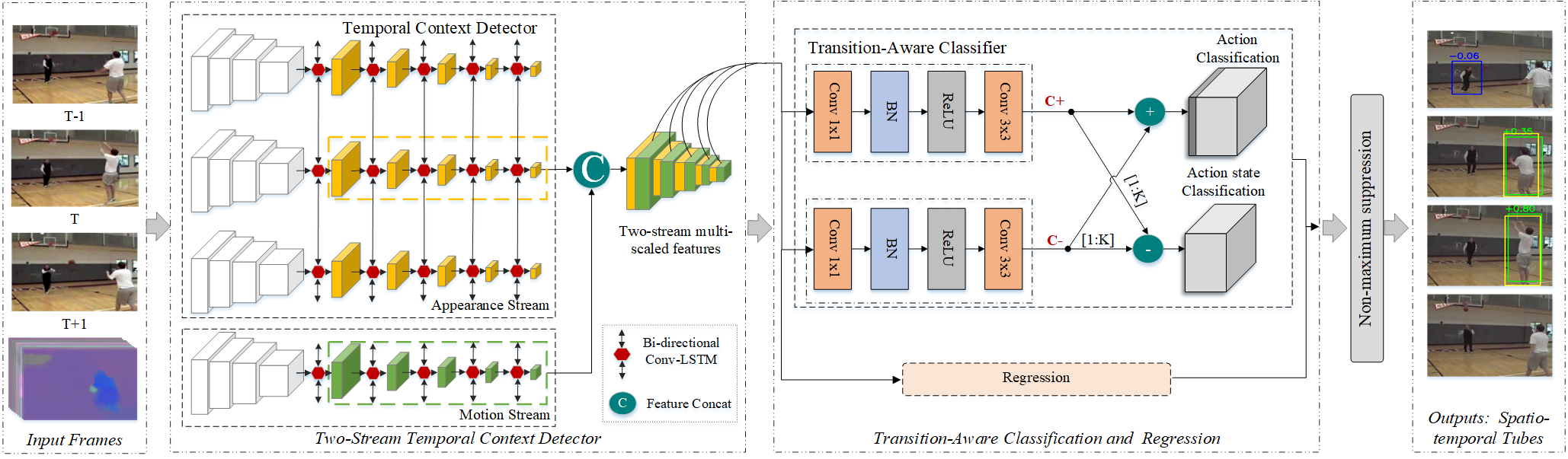}
\caption{Overall framework of the proposed TACNet. TACNet mainly contains two modules: temporal context detector and transition-aware classifier. 
In the temporal context detector, we embed several multi-scale Conv-LSTM~\cite{li2018videolstm} units in the standard SSD detector~\cite{liu_ssd:_2016} to extract temporal context. Based on the recurrent action detector, the transition-aware classifier is designed to simultaneously detect the action categories and states. Then, we can correctly localize the temporal boundaries for the target actions.
}
\label{fig:Overview}
\end{figure*}

\section{Transition-Aware Context Network}\label{theory}



\subsection{TACNet framework}

Figure~\ref{fig:Overview} illustrates the overall framework of TACNet, which mainly consists of two parts, i.e., two-stream temporal context detection and transition-aware classification \& regression. 
Although the framework is similar to most of the previous methods, especially ACT detector, temporal context detector and transition-aware classifier are proposed to significantly improve the capability of extracting long-term temporal context information and distinguishing transitional states respectively.
For the temporal context detector, we adopt two-stream SSD to construct action detection as the ACT detector does.
In addition, to extract long-term temporal context information, we embed several Bi-directional Conv-LSTM (Bi-ConvLSTM) \cite{li2018videolstm} unites into different feature maps with different scales. The Bi-directional Conv-LSTM architecture can keep the spatial layout of the feature maps, and is of benefit to performing spatial localization. 
In the transition-aware classifier, to distinguish transitional states, we propose two branches to classify the actions and action states simultaneously. Moreover, we further design a common and differential mode scheme, inspired by the basic concepts in the signal processing domain~\cite{bockelman1995combined}, to accelerate the convergence of the overall TACNet. Associating with the regression module, transition-aware classifier
can spatially detect the actions and temporally predict the temporal boundaries in the meantime. In addition, the proposed method can be embedded into various detection frameworks, and this work is based on the SSD due to its effectiveness and efficiency.

% ...

\subsection{Temporal Context Detector} \label{sec:rssd}

Long-term temporal context information is critical to spatio-temporal action detection. 
The standard SSD performs action detection from multiple feature maps with different scales in the spatial level, but it does not consider temporal context information.
To extract temporal context, we embed Bi-ConvLSTM unit into SSD framework to design a recurrent detector. 
As a kind of LSTM, ConvLSTM not only can encode long-term information, but also is more suitable to handle spatio-temporal data like video. Because the ConvLSTM unit can preserve the spatial structure of frames over time by replacing the fully connected multiplicative operations in an LSTM unit with convolutional operations.
Therefore, it is reasonable to employ ConvLSTM units into our framework to extract long-term temporal information.

In particular, we embed a Bi-ConvLSTM unit between every two layers of adjacent scales in SSD to construct the proposed temporal context detector, as shown in Figure~\ref{fig:Overview}. 
The proposed module considers the input sequences in both forward and backward directions which adopts a pair of temporal-symmetric ConvLSTM for these two directions. 
The Bi-ConvLSTM can extract a pair of features for each scale in a frame. These features are concatenated and transformed by a $1\times1$ convolutional layer to eliminate the redundancy of channels. 
By this means, the proposed temporal context detector can leverage the advantage of SSD and extract long-term temporal context information as well. 
Moreover, we also make two modifications to the Bi-ConvLSTM unit: 
(i) we replace activation function $tanh$ by $ReLU$, which can improve the performance slightly according to experimental results; 
(ii) we apply 2D dropout between inputs and hidden states to avoid over-fitting.

Compared with the ACT, our method is also efficient in terms of computational cost. 
ACT applies a sliding window with stride 1, and takes $n$ stacked frames as input for the processing of each frame.
Therefore, the computational complexity of ACT is $O(n)$.
On the contrary, we constantly process each frame twice. 
Therefore, supposing $n$ is the number of stacked frames, the computational cost of ACT and the proposed temporal context detector is $O(n)$ and $O(1)$ respectively. 
We can find that the computational cost gap increases when $n$ grows, especially $n$ can be vast when considering long-term temporal information.

\subsection{Transition-Aware Classifier}\label{sec:tam}

Proposals in the transitional state have a similar appearance to target actions and may easily confuse the detectors. 
Most of the existing methods do not provide explicit definition on these proposals but rely on post-processing algorithms to prune them or simply treat them as background. 
However, since these proposals are much different from background (e.g., scenes and other objects), treating them as background enlarges intra-class variance and limits the detection performance. 
In this paper, we propose a transition-aware classifier to perform the action category classification and the prediction of transitional state simultaneously.

\begin{figure}
\centering
\includegraphics[width=8.5cm]{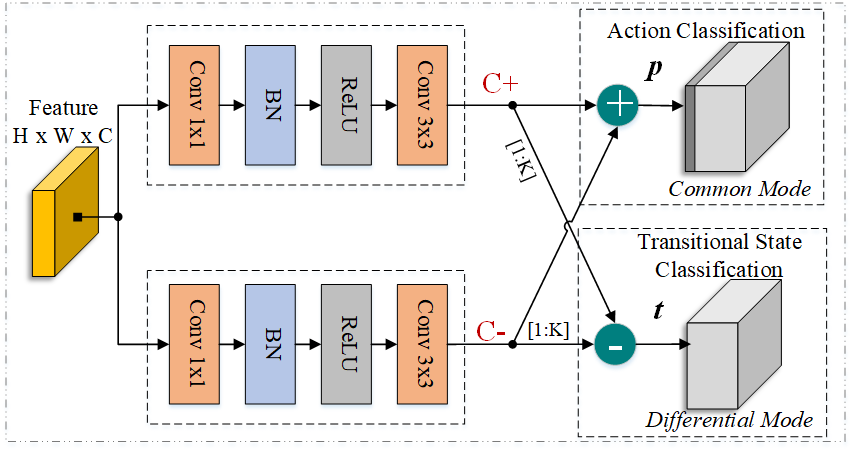}
\caption{Diagram of the transition-aware classifier. We propose a common and differential mode scheme to decouple the inter-coupled features into two branches, which predict action category and action states respectively.}
\label{fig:tam}
\end{figure}

To simultaneously predict action categories and action states, we firstly define a pair of scores, i.e. $\mathbf{c}^+=[c^+_0,c^+_1,...,c^+_K]$ and $\mathbf{c}^-=[c^-_1,c^-_2,...,c^-_K]$, where $K$ is the number of categories and $c^+_0$ is the score of background. 
The scores $\mathbf{c}^+$ and $\mathbf{c}^-$ denote the action classification scores and transitional state classification scores respectively. 
We should note that the transitional state scores have no background category. 
In the transition-aware classifier, we apply two classifiers to predict these two scores, as shown in Figure \ref{fig:tam}.
Based on these definitions, we formulate the training targets of target actions and transitional states as following:

For an active sample of category $i$, the training target should meet Eq. \ref{ra}:
\begin{equation}
c^+_i > c^-_i and\ c^+_i > c^+_j, \forall j\ne i
\label{ra}
\end{equation}
where $i,j\in[1,2,3 \dots, K]$.

While for a transitional sample of category $i$, the training target should meet Eq. \ref{ts}:
\begin{equation}
c^+_i < c^-_i.
\label{ts}
\end{equation}


In general, we can directly train TACNet based on these targets. 
However, we find that it is hard to converge when learning these two targets. 
The reason may come from that the $\mathbf{c}^+$ and $\mathbf{c}^-$ are inter-coupled, which leads to the interaction between them. 
For example, a transitional sample of the category $i$ tries to simultaneously minimize $c^+_i$ and maximize $c^-_i$ to meet Eq.\ref{ts}, which will hurt the distribution of category prediction $\mathbf{c}^+$. 

To solve this issue, we take advantage of the knowledge in signal processing domain ~\cite{bockelman1995combined} that inter-coupled signal can be decoupled by a pair of individual branches, i.e., common mode and differential mode branches.
Inspired by these concepts, we design a common and differential scheme to train our network.
In particular, the proposed transition-aware classifier still outputs $\mathbf{c}^+$ and $\mathbf{c}^-$, but the difference is that we use $\mathbf{c}^++\mathbf{c}^-$ to predict action category (upper branch in Figure \ref{fig:tam}) and $\mathbf{c}^+-\mathbf{c}^-$ to predict action state (lower branch in Figure \ref{fig:tam}). 
We formulate the decoupled targets as following:
\begin{align}
p_i &= \frac{e^{c_i^+ + c_i^-}}{\sum _{j\in [0,K]} e^{c_j^+ + c_j^-}}, \forall i\in [0,K] \label{ci}\\ 
t_i &= \frac{e^{c_i^+ - c_i^-}}{e^{c_i^+ - c_i^-} + 1}, \forall i\in [1,K].
\label{ca}
\end{align}
where \(p_i\) denotes the probability of being classified as category $i$ (Eq.\ref{ci}), which is independent to action states. 
However, we should note that \(t_i\) denotes the probability of active state but not transitional state (Eq.\ref{ca}). The probability of a transitional state is calculated as $1-t_i$.

When a sample in transitional state tries to minimize $\mathbf{c}^+-\mathbf{c}^-$, we suppose that $\mathbf{c}^+$ changes to $\mathbf{c}^+-\boldsymbol{\lambda}$, and $\mathbf{c}^-$ will change to $\mathbf{c}^-+\boldsymbol{\lambda}$ (since these two branches share the same gradient magnitude). 
The category prediction $\mathbf{c}^++\mathbf{c}^-$ will have no change. 
By this means, the predictions of the action category and action state are decoupled. 
In our experiments, we find that the network can converge easily and the prediction has a very slight effect on each other. 



\subsection{Training and Inference Procedures}\label{sec:training_inference}


In the training phase of TACNet, we denote \(P\) as the positive samples which have more than 0.5 IoU with at least one ground-truth bounding box. 
We apply $T$ and $\mathbf{v}$ to denote the positive sample set of transitional states and the corresponding predicted action categories respectively. 
However, there are no annotations for transitional states in the existing untrimmed datasets so far. 
According to the definition of transitional states, we propose a ``simple-mining'' strategy to label the positive samples of the transition states. 
In detail, the detected boxes are treated as transitional state samples when their scores meet $c^+_i>c^+_0$, but their corresponding frames have no ground-truth annotations. These transitional state samples are further applied to train TACNet. 

We employ the same loss function as SSD for regression \(\mathcal{L}_{\text{reg}}\). 
Besides, we introduce the classification loss \(\mathcal{L}_{\text{cls}}\) and transitional loss \(\mathcal{L}_{\text{trans}}\) based on the classification scores \(\mathbf{p}\) and action state scores \(\mathbf{t}\) illustrated in Eq.\ref{eq:trans}:

\begin{equation}
\begin{aligned}
\label{eq:trans} 
&\mathcal{L}_{\text{cls}}=-\sum _{j\in P} \log  p^j_y-\sum _{j\in G\backslash P} \log  p^j_0, \\
&\mathcal{L}_{\text{trans}}=-\sum _{j\in P} \log  t^j_y-\sum _{j\in T} \log  \left(1-t^j_{\mathbf{v}_j}\right),\\
&\mathbf{v}_j=\underset{i\in [1,K]}{\text{arg max}}(p_i^j), 
T=\left\{j\left|{\mathbf{v}_j}>0\right.,j\in U\backslash G\right\},
\end{aligned} 
\end{equation}
where $U$ and $G$ refers to the set of all available anchors and anchors in the images which have groudtruth annotations respectively, $p^j_i$ is the predicted probability of $j$-th anchor with category $i$, and $t^j_i$ is the probability of $j$-th transitional anchor with predicted category $i$.

We optimize the proposed TACNet with the combined loss:
\begin{equation}\label{eq:loss} 
\begin{split}
\mathcal{L}&=\frac{1}{N_p}(\mathcal{L}_{\text{cls}} + \mathcal{L}_{\text{reg}}) + \frac{1}{N_t}\mathcal{L}_{\text{trans}}, \\ 
\end{split} 
\end{equation} 
where \(N_p\) and \(N_t\) denote the number of positive samples $P$ and transitional samples $T$ respectively. 
We train each stream of our network in an end-to-end manner. The experiments demonstrate that the branches of classification and transition can be optimized simultaneously.

In the inference phase, TACNet takes video clips as input and output three items: spatial detection boxes, classification scores, and action state scores. 
To construct spatio-temporal action tubes, we first apply the greedy algorithm~\cite{singh2017online} on the frame-level detections using category scores to construct candidate tubes. 
Secondly, we apply action state predictions to perform temporal detection. 
In the experiments, we find that the action state scores are discontinuous, and hence apply the watershed segmentation algorithm~\cite{zhao2017temporal} to trim candidate tubes for temporal detection.  
Besides, inspired by ACT~\cite{kalogeiton2017action}, we introduce a micro-tube refinement (MR) procedure which constructs a micro-tube for proposals in the same spatial position of adjacent frames to average the predicted category scores. The score of each box is set to the product of category score and action state score.

\section{Experiments}
\subsection{Experimental setup}
We evaluate TACNet on two datasets: UCF101-24~\cite{soomro2012ucf101} and J-HMDB~\cite{jhuang2013towards}. 
The UCF101-24 dataset contains 3207 videos for 24 classes. Approximate 25\% of all the videos are untrimmed. 
The J-HMDB dataset contains 928 videos with 33183 frames for 21 classes. 
Videos in this dataset are trimmed to actions. 
Hence we only use it to evaluate the spatial detection of the proposed TACNet.

We apply the metrics of \textit{frame-mAP} and \textit{video-mAP} \cite{gkioxari2015finding} to evaluate TACNet at frame-level and video-level respectively. 
The frame-mAP and video-mAP measure the area under the precision-recall curve of detections for each frame and action tubes respectively.
Therefore, frame-mAP measures the ability of classification and spatial detection in a single frame, and video-mAP can also evaluate the performance of temporal detection.
The prediction is correct when its overlap with ground-truth box/tube above a certain threshold and predicted category is correct.
In this paper, we apply constant IoU threshold (0.5) to evaluate frame-mAP and variable IoU thresholds (i.e 0.2, 0.5, 0.75 and 0.5:0.95) to evaluate video-mAP.

We provide some implementation details in TACNet as follows.
Input frames are resized to 300x300. 
The clip size \(L\) is set as 16 for training and inference. 
The number of flow images \(S\) is set as 5 for each frame.
The probability of 2D dropout is set as 0.3. 
In the training phase, we stack 32 clips as a mini-batch and apply data augmentation of color jitter, cropping, rescaling, and horizontal flipping. 
We use the hard-negative-mining strategy that only the hardest negatives up to the same quantity of positives are kept to calculate the loss.
Following the previous works, we separate training the appearance stream from the motion stream.
To train the appearance branch, we set the learning rate as 0.01 for initial learning and decrease it by 10 times in every 20 epochs. 
We employ a warmup scheme \cite{goyal2017accurate} with the learning rate of 0.001 in the training phase. To train the motion branch, we take the parameters from the appearance branch as the initial parameters and set the initial learning rate as 0.001. Furthermore, we fine-tune the fusion network with a consistent learning rate of 0.0001.
Besides, we optimize TACNet by stochastic gradient descent (SGD) with momentum of 0.9.
In the inference phase, the number of micro-tube frames is set as 8 for micro-tube refinement procedure.

\begin{table}[!t]
\begin{threeparttable}
 \renewcommand{\arraystretch}{1.3}
 \footnotesize
 \caption{Performance Comparison on J-HMDB.}
  \begin{center}
  \label{table:compare_jhmdb}
  \begin{tabular}{p{1.8cm}|p{1.5cm}<{\centering}|cccc}
    \hline
    \multirow{2}{*}{Method} 	& \multirow{2}{*}{Frame-mAP} & \multicolumn{4}{c}{Video-mAP} \\ 
    \cline{3-6} &  & 0.2 & 0.5 & 0.75 & 0.5:0.95 \\ 
    \hline
    SSD & 49.5 & 60.9 & 60.1 & 41.5 & 33.8 \\
    TCD\tnote{1} & 54.1 & 65.0 & 64.5 & 45.3 & 35.1 \\
    TS\tnote{2} +SSD & 56.4 & 70.9 & 70.3 & 48.3 & 42.2 \\
    TS+TCD & 61.5 & 74.1 & 73.4 & 52.5 & 44.8 \\
    \hline
    \bf{TS+TCD+MR\tnote{3}} & \bf{65.5} & \bf{74.1} & \bf{73.4} & \bf{52.5} & \bf{44.8} \\
    \hline
  \end{tabular}
  \begin{tablenotes}
\item[1] TCD: Temporal Context Detector;  
\item[2] TS: Two-Stream;
\item[3] MR: Micro-tube refinement.
\end{tablenotes}
  \end{center}
 \end{threeparttable}
 \vspace{-3mm}
\end{table}

\subsection{Analysis of Temporal Context Detector}

We perform several experiments to evaluate the effectiveness of the proposed temporal context detector under different configurations on J-HMDB and UCF101-24 dataset. 
We report the frame-level and video-level performance in Table \ref{table:compare_jhmdb} - \ref{table:compare_ucf24}.


\begin{table}[!t]
\renewcommand{\arraystretch}{1.3}
\begin{threeparttable}
\footnotesize
\caption{Performance Comparison on UCF101-24.}
\begin{center}
\label{table:compare_ucf24}
\begin{tabular}{l|p{1cm}<{\centering}|cccc}
\hline
\multirow{2}{*}{Method} & \multirow{2}{*}{F-mAP} & \multicolumn{4}{c}{Video-mAP} \\ \cline{3-6} &  & 0.2 & 0.5 & 0.75 & 0.5:0.95 \\ 
\hline
SSD & 65.3 & 69.1 & 39.3 & 16.3 & 18.4 \\
TCD & 67.5 & 72.2 & 45.0 & 17.4 & 19.4 \\
TS+SSD	& 66.5 & 74.3 & 47.5 & 19.2 & 21.0 \\
TS+TCD & 68.7 & 77.3 & 50.6 & 20.9 & 23.0 \\
TS+TCD+BG\tnote{1} & 68.3 & 77.0 & 48.9 & 20.8 & 22.4 \\
TS+SSD+TAC\tnote{2} & 67.1 & 74.5 & 49.0 & 20.1 & 21.8 \\
TS+TCD+TAC & 69.7 & 77.5 & 52.9 & 21.8 & 24.1 \\
\hline
\bf{TS+TCD+TAC+MR} & \bf{72.1} & \bf{77.5} & \bf{52.9} & \bf{21.8} & \bf{24.1} \\
\hline
\end{tabular}
\begin{tablenotes}
\item[1] BG: This method simply treats transitional states as background, and can be treat as hard-mining method;
\item[2] TAC: Transition-Aware Classifier;
\end{tablenotes}
\end{center}
\end{threeparttable}
\vspace{-3mm}
\end{table}



From the results, we can find that temporal context detector significantly outperforms standard SSD on both datasets. 
On the dataset of J-HMDB, the proposed temporal context detector with a two-stream configuration obtains the improvements of 5.1\% and 3.1\% the in terms of frame-mAP and video-mAP (IoU threshold is 0.5) respectively. 
The improvements on UCF101-24 are 2.2\% and 3.1\% respectively. 
These results clearly demonstrate that temporal context information can effectively improve performance.

\begin{figure*}[h]
\centering
\includegraphics[width=\textwidth]{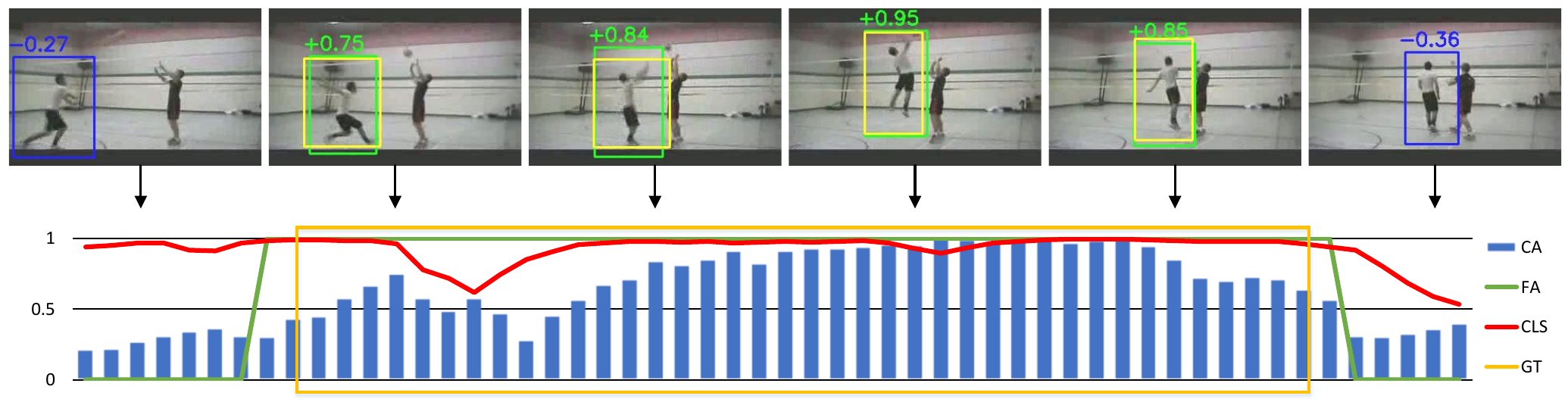}
\caption{Visualized analysis of transition-aware classifier when taking ``Basketball pitch'' action as an example.
Top row: detection boxes with their corresponding action state scores; bottom row: the procedure with different predictions including Coarse action state score (CA), Refined action state score (FA), Classification score (CLS), and compared to Ground-truth (GT). 
Based on these predictions, we temporally trim the detection whose FA score is larger than $0.5$, while the others are treated as transitional samples.
}
\label{fig:transition_result}
\end{figure*}

\subsection{Analysis of Transition-Aware Classifier}

We perform several experiments to evaluate the transition-aware classifier on the untrimmed UCF101-24 dataset. 
We report the experimental results with different configurations in Table~\ref{table:compare_ucf24}, and the per-class performance of untrimmed categories in Table \ref{table:category_ucf24}. 
We present some visualized analysis of transition-aware classifier in Figure \ref{fig:transition_result}.


In Table \ref{table:compare_ucf24}, the results of improvement prove its effectiveness of the proposed transition-aware classifier with different settings.
It should be noted that the hard-mining method (``BG'' in the table) leads to a slight performance drop, which means that simply treating transition sates as the background is unreasonable. 
In contrast, the transition-aware classifier achieves significant improvement in terms of both metrics.
In particular, it improvements performance by 2.3\% in terms of video-mAP when IoU is 0.5. 
Therefore, the results can clearly demonstrate that it is crucial to define transitional states and the proposed transition-aware classifier can well distinguish these states.

In Table \ref{table:category_ucf24}, we can find that transition-aware classifier can obtain obvious improvement in the untrimmed videos. 
Especially, we outperform the baseline without a transitional-aware classifier and ACT by 1.7\% and 7.8\% in terms of video-mAP respectively. 
The transition-aware classifier achieves 8\% improvement in terms of temporal detection when only considering temporal IoU (``Temporal'' in the table),. 
Therefore, these results demonstrate the capability of a transition-aware classifier for temporal extent detection.

\begin{figure*}[h]
\centering
\includegraphics[width=\textwidth]{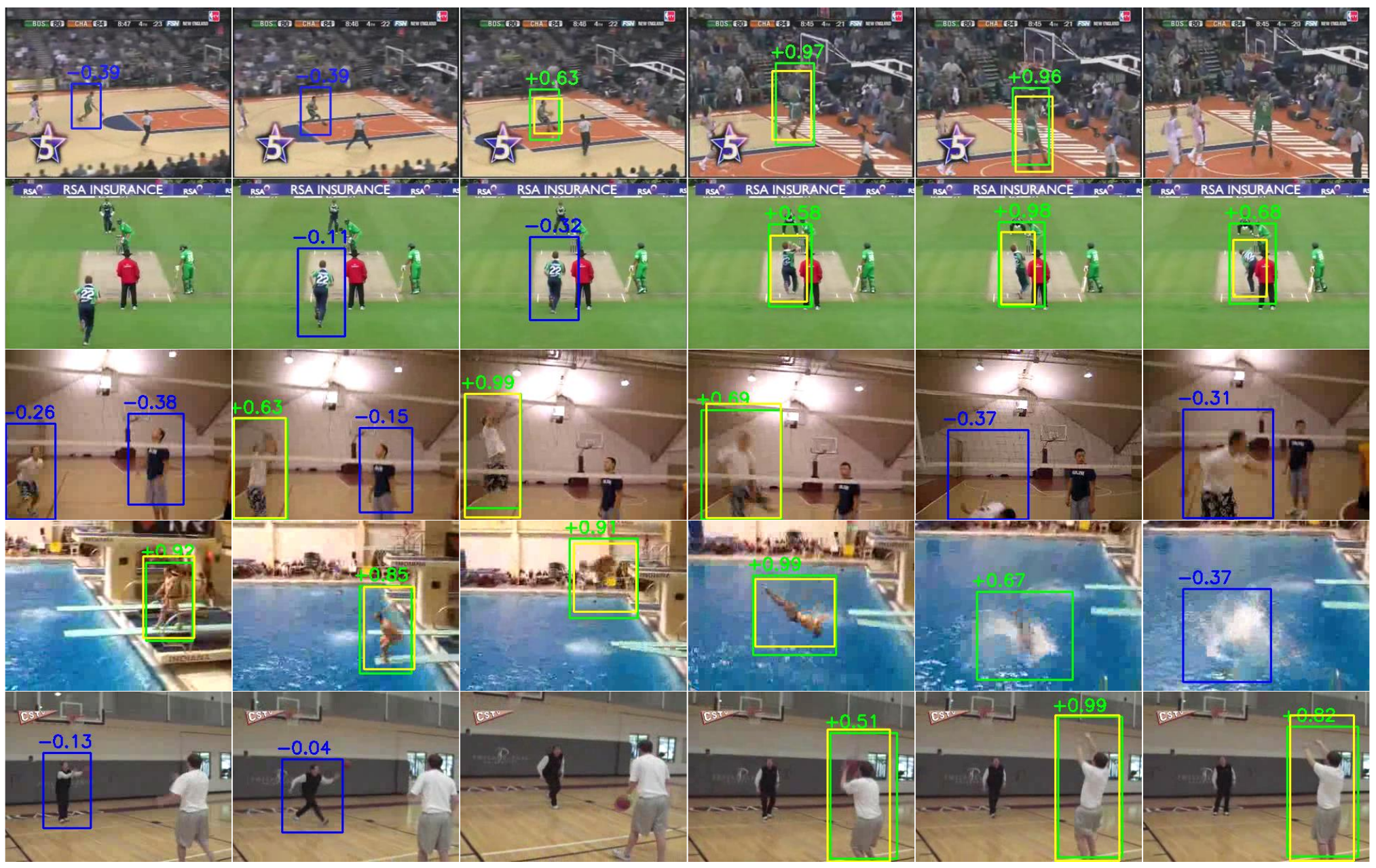}
\caption{
Predicted results of the proposed TACNet in four states: (a) Background (w/o box), (b) Transitional state (blue box), (c) Active state (green box) and (d) Ground-truth (yellow box). }
\label{fig:result}
\end{figure*}

In Figure \ref{fig:transition_result}, we take a ``Volleyball'' action instance as an example to intuitively show the reason for the performance improvement of a transition-aware classifier. 
We can find that it is difficult to distinguish transitional states by only considering classification scores. 
However, action state scores can help to trim the actions well temporally. 
More results can be found in Figure \ref{fig:result}. 
In Figure \ref{fig:result}, we can find that the proposed classifier can also distinguish the action states for multiple instances.


\begin{table*}[!htbp]
\renewcommand{\arraystretch}{1.3}
\footnotesize
\footnotesize
\begin{threeparttable}
\begin{center}
\caption{Comparison with the state-of-the-art on J-HMDB (trimmed) and UCF101 (untrimmed)}
\label{table:state}
\begin{tabular}{c|c|cccc|c|cccc|c|c}
\hline
\multirow{3}{*}{Method} &\multicolumn{5}{c|}{J-HMDB}  								  &\multicolumn{5}{c|}{UCF101-24 (Full)\tnote{1}} &\multicolumn{2}{c}{UCF101-24 (Untrimmed)\tnote{1}}  \\
\cline{2-13}
						& \multirow{2}{*}{F-mAP} & \multicolumn{4}{c|}{Video-mAP}  & \multirow{2}{*}{F-mAP}   & \multicolumn{4}{c|}{Video-mAP} & \multirow{2}{*}{F-mAP} & \multicolumn{1}{c}{Video-mAP} \\
\cline{3-6} 
\cline{8-11}
\cline{13-13} 
& & 0.2 & 0.5 & 0.75 & 0.5:0.95 &  & 0.2 & 0.5 & 0.75 & 0.5:0.95 &  & 0.5 \\
\hline
Saha~\cite{saha2016deep} 	& - & 72.6 & 71.5 & 43.3 & 40.04  							  & - & 66.7 & 35.9 & 7.9 & 14.4 & - & - \\
Peng~\cite{peng2016multi} 	& 58.5 & 74.3 & 73.1 & - & -  & 65.7 & 73.5 & 32.1 & 2.7 & 7.3 & - & - \\
Singh~\cite{singh2017online} & - & 73.8 & 72.0 & 44.5 & 41.6 & - & 73.5 & 46.3 & 15.0 & 20.4 & - & - \\
Hou~\cite{hou2017tube} & 61.3 & \bf{78.4} & \bf{76.9} & - & -  & 41.4 & 47.1 & - & - & - & - & - \\
Becattini~\cite{becattini2017done} & - & - & - & - & - 	& - & 67.0 & 35.7 & - & - & - & - \\
Kalogeiton~\cite{kalogeiton2017action} & \bf{65.7} & 74.2 & 73.7 & 52.1 & 44.8 & 69.5 & 76.5 & 49.2 & 19.7 & 23.4 & 52.1 & 23.5 \\
\hline
Ours & 65.5 & 74.1 & 73.4 & \bf{52.5} & \bf{44.8} & \bf{72.1} & \bf{77.5} & \bf{52.9} & \bf{21.8} & \bf{24.1} & \bf{58.0} & \bf{31.3} \\
\hline
\end{tabular}
\begin{tablenotes}
\item[1] UCF101-24 is a mixture dataset which is made up of untrimmed categories and trimmed categories, thus we evaluate our approaches in two criterions to  fully illustrate the performance gain on untrimmed videos.
\end{tablenotes}
\end{center}
\end{threeparttable}
\end{table*}

\begin{table}[H]
\renewcommand{\arraystretch}{1.1}
\footnotesize
\begin{threeparttable}
\centering
\caption{Per-class performance of untrimmed categories on UCF101-24.}
\label{table:category_ucf24}
\begin{tabular}{p{1.8cm}<{\centering}|p{0.4cm}<{\centering}p{0.4cm}<{\centering}|p{0.4cm}<{\centering}p{0.4cm}<{\centering}p{0.4cm}<{\centering}p{0.4cm}<{\centering}p{0.4cm}<{\centering}}
\hline
\multirow{3}{*}{Category} & \multicolumn{2}{c|}{\multirow{2}{*}{Frame-mAP}} & \multicolumn{5}{c}{Video-mAP} \\ \cline{4-8} 
 & \multicolumn{2}{c|}{} & \multicolumn{3}{c|}{Spatio-temporal\tnote{1}} & \multicolumn{2}{c}{Temporal\tnote{2}} \\ \cline{2-8} 
 & Tw\tnote{3} & To\tnote{4} & Tw & To & \multicolumn{1}{c|}{ACT} & Tw & To \\ \hline
Basketball & 44.0 & 34.8 & 5.5 & 0.3 & \multicolumn{1}{c|}{0.0} & 25.8 & 9.3 \\
Dunk & 57.1 & 53.0 & 18.9 & 5.3 & \multicolumn{1}{c|}{1.2} & 88.2 & 76.6 \\
CliffDiving & 74.9 & 74.4 & 42.9 & 45 & \multicolumn{1}{c|}{39.9} & 84.4 & 84.4 \\
CricketBowl & 39.7 & 42.4 & 3.6 & 0.9 & \multicolumn{1}{c|}{1.1} & 25.5 & 14.0 \\
Diving & 85.9 & 82.3 & 52.1 & 43.2 & \multicolumn{1}{c|}{26.1} & 84.7 & 84.7 \\
GolfSwing & 58.9 & 55.8 & 65.9 & 49.5 & \multicolumn{1}{c|}{51.0} & 70.4 & 70.4 \\
LongJump & 58.6 & 59.0 & 50.9 & 46.2 & \multicolumn{1}{c|}{71.1} & 68.0 & 66.2 \\
PoleVault & 64.0 & 63.8 & 57.2 & 60.7 & \multicolumn{1}{c|}{44.6} & 77.4 & 77.4 \\
TennisSwing & 46.4 & 40.2 & 2.2 & 0.1 & \multicolumn{1}{c|}{0.5} & 9.6 & 8.0 \\
Volleyball & 50.6 & 44.1 & 13.4 & 0.9 & \multicolumn{1}{c|}{0.0} & 48.8 & 12.0 \\ \hline
Mean-AP & \bf{58.0} & 55.0 & \bf{31.3} & 25.2 & \multicolumn{1}{c|}{23.5} & \bf{58.3} & 50.3 \\ \hline
\end{tabular}
\begin{tablenotes}
\item[1] Spatio-temporal: evaluating the performance in terms of standard video-mAP;
\item[2] Temporal: evaluating the performance again ground-truth by only considering temporal IoU while without spatial IoU;
\item[3] Tw: results with transition-ware classifier;
\item[4] To: results without a transition-aware classifier.
\end{tablenotes}
\end{threeparttable}
\end{table}

\begin{table}[H]
\renewcommand{\arraystretch}{1.3}
\footnotesize
  \caption{Performance of different detector and base model on UCF101-24 dataset without micro-tube refinement.}
  \begin{center}
  \label{table:different_model}
  \begin{tabular}{p{0.9cm}|p{1cm}<{\centering}|p{1.5cm}<{\centering}|p{0.4cm}<{\centering}p{0.4cm}<{\centering}p{0.4cm}<{\centering}p{0.8cm}<{\centering}}
    \hline
    \multirow{2}{*}{Method}  &\multirow{2}{*}{Model}	& \multirow{2}{*}{Frame-mAP} & \multicolumn{4}{c}{Video-mAP} \\ 
    \cline{4-7}
    								& &  & 0.2 & 0.5 & 0.75 & 0.5:0.95 \\ 
    \hline
    SSD   & VGG16					& 69.7 & 77.5 & 52.9 & 21.8 & 24.1 \\
    DSSD   & VGG16					& 70.1 & 77.5 & 53.0  & 22.1 & 24.5 \\
    SSD   & Resnet50					& 72.0 & 78.9 & 54.4  & 23.0 & 24.8 \\
    \bf{DSSD} & Resnet50				& \bf{74.6} & \bf{79.2} & \bf{54.6} & \bf{23.3} & \bf{25.4} \\
    \hline
  \end{tabular}
  \end{center}
  \vspace{-3mm}
\end{table}

\subsection{Exploration on advanced backbone}
In this section, we explore different detectors and models using our method. 
We respectively replace the detector and base model with Deconvlution-SSD~\cite{fu_dssd:_2017} and Resnet-50 respectively, and show the results in Table \ref{table:different_model}. 
All the models are pretrained on the ImageNet. Two streams, temporal context detector, and transition-aware classifier are also applied. 
We can find that the employment of advanced models can further improve performance.

\subsection{Comparison with the state-of-the-art}
We compare TACNet with state-of-the-art methods in terms of both frame-mAP and video-mAP, and the results are shown in Table \ref{table:state}.
From the results, we see that TACNet outperforms all these previous methods on temporally untrimmed UCF101-24 dataset in terms of both metrics. 
In particular, we surpass ACT, which is the current state-of-the-art, by 3.7\% in terms of video-level when the IoU threshold is 0.5.  
On the trimmed J-HMDB dataset, ACT and T-CNN outperform TACNet in terms of frame-mAP and video-mAP respectively. 
We think there are two reasons: 
(i) these two methods directly generate action tubes, which are suitable for the trimmed dataset;
(ii) J-HMDB dataset is relatively simple, especially for localization since only one instance in each frame. 
However, for action detection, video-mAP is more suitable to evaluate the performance than frame-mAP in both spatial and temporal domain, and in terms of video-mAP TACNet obtains a competitive performance with ACT.
On the more challenging untrimmed UCF101-24 dataset, we can find that TACNet significantly outperforms T-CNN and ACT.
It even improves the performance against T-CNN by 28.5\% and 30.4\% in terms of frame-mAP and video-mAP when IoU is 0.5. 
Therefore, the superior performance of TACNet demonstrates the importance of long-term temporal context information and transitional state detection.

\section{Conclusions}

This paper aims to improve the performance of action detection.
In particular, we find that it is critical to extract long-term temporal context information and distinguish transitional states. 
Based on these observations, we propose a TACNet which consists of a temporal context detector and a transitional-aware classifier
We extensively explore TACNet on two public datasets. 
From the experimental results, we find TACNet can significantly improve the performance and surpass the state-of-the-art on the challenging untrimmed dataset. 
The performance improvements of TACNet come from both temporal detection and transition-aware method.
In future work, we will continue our exploration on how to further improve temporal detection by considering the relation between actors and their surrounding peoples or objects.


\section{Acknowledgment}
This research was supported by the National Key R\&D Program of China (No. 2017YFA0700800).

{\small
\bibliographystyle{ieee}
\bibliography{Reference/Reference}
}

\end{document}